\documentclass[sigconf, nonacm]{acmart}
\usepackage{booktabs}  
\usepackage{graphicx}  
\usepackage{array}     
\usepackage{multirow}  
\usepackage{xcolor}    
\usepackage{tcolorbox}
\usepackage{tikz}
\usepackage{tabularx}
\usepackage{enumitem}

\newcommand\vldbyear{2025}
\newcommand\vldbworkshop{Tabular Data Analysis (TaDA)}
\newcommand\vldbauthors{\authors}
\newcommand\vldbtitle{\shorttitle} 
\newcommand\vldbavailabilityurl{URL_TO_YOUR_ARTIFACTS}
\newcommand\vldbpagestyle{plain}

\begin{document}
\title{StructText: A Synthetic Table-to-Text Approach for \\ Benchmark Generation with Multi-Dimensional Evaluation}

\author{Satyananda Kashyap}
\affiliation{%
  \institution{IBM Research}
}
\email{satyananda.kashyap@ibm.com}

\author{Sola Shirai}
\affiliation{%
  \institution{IBM Research}
}
\email{solashirai@ibm.com}

\author{Nandana Mihindukulasooriya}
\affiliation{%
  \institution{IBM Research}
}
\email{nandana@ibm.com}

\author{Horst Samulowitz}
\affiliation{
  \institution{IBM Research}
}
\email{samulowitz@us.ibm.com}

\begin{abstract}
Extracting structured information from text, such as key-value pairs that could augment tabular data, is quite useful in many enterprise use cases. Although large language models (LLMs) have enabled numerous automated pipelines for converting natural language into structured formats, there is still a lack of benchmarks for evaluating their extraction quality, especially in specific domains or focused documents specific to a given organization. Building such benchmarks by manual annotations is labour-intensive and limits the size and scalability of the benchmarks.

In this work, we present \texttt{StructText}, an end-to-end framework for automatically generating high-fidelity benchmarks for key-value extraction from text using existing tabular data. It uses available tabular data as structured ground truth, and follows a two-stage ``plan-then-execute'' pipeline to synthetically generate corresponding natural-language text. To ensure alignment between text and structured source, we introduce a multi-dimensional evaluation strategy that combines (a) LLM-based judgments on factuality, hallucination, and coherence and (b) objective extraction metrics measuring numeric and temporal accuracy. 

We evaluated the proposed method on 71,539 examples across 49 datasets. Results reveal that while LLMs achieve strong factual accuracy and avoid hallucination, they struggle with narrative coherence in producing extractable text. Notably, models presume numerical and temporal information with high fidelity yet this information becomes embedded in narratives that resist automated extraction.

We release a framework, including datasets, evaluation tools, and baseline extraction systems, to support continued research. Our findings highlight a critical gap: Models can generate accurate text but struggle to maintain information accessibility, a key requirement for practical deployment in different sectors and demanding both accuracy and machine processability.
\end{abstract}

\maketitle

\pagestyle{\vldbpagestyle}
\begingroup\small\noindent\raggedright\textbf{VLDB Workshop Reference Format:}\\
\vldbauthors. \vldbtitle. VLDB \vldbyear\ Workshop: \vldbworkshop.\\ 
\endgroup
\begingroup
\renewcommand\thefootnote{}\footnote{\noindent
This work is licensed under the Creative Commons BY-NC-ND 4.0 International License. Visit \url{https://creativecommons.org/licenses/by-nc-nd/4.0/} to view a copy of this license. For any use beyond those covered by this license, obtain permission by emailing \href{mailto:info@vldb.org}{info@vldb.org}. Copyright is held by the owner/author(s). Publication rights licensed to the VLDB Endowment. \\
\raggedright Proceedings of the VLDB Endowment. 
ISSN 2150-8097. \\
}\addtocounter{footnote}{-1}\endgroup

\ifdefempty{\vldbavailabilityurl}{}{
\vspace{.3cm}
\begingroup\small\noindent\raggedright\textbf{VLDB Workshop Artifact Availability:}\\
The source code, data, and/or other artifacts have been made available at \url{https://huggingface.co/datasets/ibm-research/struct-text}\\
\url{https://github.com/ibm/struct-text}.
\endgroup
}

\section{Introduction} 
Large language models (LLMs) have unlocked new opportunities in automating knowledge extraction and generation tasks~\cite{kasner-dusek-2024-beyond} across structured and unstructured data. Recent advances in model capabilities enable quality reporting from various sources. Industries increasingly automate and adopt LLM powered workflows for converting databases, sensor feeds and other heterogeneous structured records into human readable insights~\cite{app14062506, qian2024evolutionllmadoptionindustry}.

One particularly promising use case for LLM-powered knowledge extraction is to perform text-to-table extraction~\cite{wu2022texttotablenewwayinformation, deng2024texttupletableinformationintegrationtexttotable}, where free form text is processed into tabular data formats. Converting natural language text into tabular data can allow it to more easily integrate with existing tabular data as well as leverage the ecosystem of mature technologies surrounding efficient indexing and querying over tables. This in turn can enable developers and users to make use of text data more effectively and precisely -- for example, rather than relying on full text search or embedding-based similarity metrics to retrieve potentially useful documents to answer a question, we can instead convert the documents to tables and query even large amounts of data effectively using SQL.

However, research surrounding the topic of text-to-table extraction is still constrained by the absence of \emph{broad, domain-spanning} benchmarks and evaluations. Classic datasets such as
WebNLG~\cite{gardent-etal-2017-webnlg}, WikiBio~\cite{lebret-etal-2016-neural},
ToTTo~\cite{parikh-etal-2020-totto}, LogicNLG~\cite{chen-etal-2020-logic2text} all target a single content domain
(typically Wikipedia) and judge mainly surface-form overlap, leaving numerical fidelity and schema coverage untested.  Synthetic pipelines like SynthIE~\cite{josifoski-etal-2023-exploiting} show that LLMs can bootstrap data, but their
evaluations likewise remain confined to news or encyclopaedic prose. Industry applications require verifiable and high quality information transfer, especially in highly regulated areas such as the finance, healthcare, and legal domains ~\cite{wu2023bloomberggptlargelanguagemodel, hardianhealth_regulatory_llms, chaudhary2025llmfactuality}. While some existing work has explored the quality of LLMs in data-to-text workflows ~\cite{kasner-dusek-2024-beyond}, most existing systems either focus on structured-to-structured pipelines, such as SQL querying and table QA, or on unstructured-to-unstructured tasks like summarization and open-ended generation. Existing structured to text benchmarks tend to focus on narrow domains rather than mapping real-world complexity. Furthermore, we lack the deeper level evaluations that measure information preservation or factual accuracy creating a disconnect between research benchmarks and practical needs.

To facilitate the text-to-table extraction research and evaluation, we introduce \texttt{StructText}, an automated workflow to generate text, extract tables, and evaluate the quality of both structured and unstructured data. Our workflow operates over any existing tabular data, processing it to automatically generate natural language text from table rows, and subsequently evaluate the quality of structured extraction over the synthetically generated text. By generating unstructured text starting from a tabular data source, StructText enables the creation of new datasets to benchmark the quality of text-to-table extraction for any domain, using the original data source as the ground-truth for evaluation.

As demonstrated by prior work such as SynthIE ~\cite{josifoski-etal-2023-exploiting}, there is an inherent asymmetry in difficulty for LLMs: generating fluent text from structured data inputs seems to be easier for LLMs than extracting structured information from natural language text. We leverage this asymmetry by first synthesizing natural-language passages from tabular data, then using these synthetic text to rigorously evaluate and advance LLM-based information-extraction capabilities.

\subsection{Our contributions}
This work makes three primary contributions to text-to-table generation benchmarking:

\textbf{Synthetic text and benchmark generation}: We present a methodology to automatically process existing tabular data to produce new benchmarking datasets. Given any tabular data source, we leverage LLMs to select meaningful groupings of columns to produce synthetic text for each row. Our method minimizes the need for human effort while enabling the creation of new text-to-table benchmark datasets for any domain.

\textbf{Multi-dimensional evaluation framework}: We devise multi-dimensional evaluation criteria for the generated synthetic text. The evaluation measures generation across four critical dimensions: factuality, hallucination, coherence, and numeric/temporal accuracy. Our framework incorporates LLM-as-judge methodology, specifically adapted for structural data scenarios combined with specialized hybrid models for numerical and temporal reasoning. This evaluation approach measures if the generated text is faithful~\cite{li2022faithfulness} to the original structured data that is being used as input, \textit{i.e.}, whether generated text truly preserves source information so that it is possible for a system to extract the same information as in the ground truth. Furthermore, it ensures that the generated text is of high quality and coherent in addition to being factually correct. 

\textbf{Benchmark dataset}: To support the feasibility of our approach, we produce a large-scale benchmarking dataset spanning practical domains including finance, healthcare, and diverse sources of information originating from Wikidata. Unlike existing simplified academic scenarios, it reflects real-world complexity found in regulated industries. Using this initial dataset, we perform baseline experiments and evaluation surrounding the benchmark creation and text-to-table extraction tasks.

\textbf{Baseline extraction method}:
We implement a text-to-table extraction pipeline that converts a collection of natural-language passages into tabular format, with each piece of text corresponding to a single row with one or more columns and values.  To quantify performance, we define a set of evaluation metrics guided by the existing work on key-value pair extraction from text. The baseline acts as a reference for the more advanced approaches to be compared, and also as a validation level of difficulty in the task, \textit{i.e.}, that the generated synthetic text is nontrivial for modern LLMs to parse back into structured data.

\section{Related Work}

Early approaches on evaluating table-to-text methodologies relied on template-based systems with rule-based evaluation focusing on grammatical correctness. In the neural era, BLEU~\cite{bleu} and ROUGE~\cite{rouge} metrics borrowed from machine translation inadequately captured factual accuracies. Recent LLM as a judge methods show promise but lack standardized application to table-to-text benchmarks where the information fidelity is critical. Our approach draws from and contributes to several strands of recent research in LLM evaluation, factuality detection, and structured generation.

\textbf{Key-value extraction benchmarks:} Early key-value pair extraction benchmarks were focused on specific semi-structured documents such as receipts (e.g., CORD~\cite{park2019cord}) with a pre-defined small key set. KVP10K~\cite{NaparstekASAYBPRDPFALLN24} dataset formulates this task as an open vocabulary extraction task without a predefined key set. Typically, creating such benchmarks involves a significant amount of effort from domain experts for manual annotations. In this paper, we propose a method to automatically generate benchmarks that can be used to test key-value pair extraction approaches using tabular data from any given domain.

\textbf{Benchmarks for Structured-to-Text:} Systems like ToTTo~\cite{parikh-etal-2020-totto}, LogicNLG~\cite{chen-etal-2020-logic2text}, WebNLG~\cite{gardent-etal-2017-webnlg}, WikiBio~\cite{lebret-etal-2016-neural} evaluate table-to-text generation, but typically in isolation from downstream tasks. They are limited to simplified toy problems, repetitive text patterns.  Unlike these datasets, our benchmark defines a comprehensive evaluation that focused on multi-domain complexity spanning SEC Filings~\cite{sec} and a diverse fields in WikiDB, highlighting real world data complexity with information fidelity assessment through multi-stage evaluations instead of generation only metrics. 

\textbf{LLM-as-a-Judge Evaluation:} Recent work demonstrates that LLMs can perform evaluative tasks traditionally done by humans. MT-Bench/Chatbot Arena~\cite{zheng-mtbench2020} showed high agreement with human preferences similar to human-human agreement. Similarly G-Eval~\cite{liu2023geval} and TrueTeacher~\cite{gekhman-etal-2023-trueteacher} show strong correlations between LLM and human judgments across dimensions like factuality and coherence. FACTS~\cite{jacovi2025facts} extends this line to long-document factuality by using grounded context-aware scoring. General LLMs as a judge focus on open-ended text quality and we build on these frameworks by adapting them specifically to evaluate structured to text conversion. Our design focuses on a three dimensional rubric covering factuality, hallucination and coherence tailored for structured data.

\textbf{Evaluation Criteria:} Prior benchmarks often focus narrowly—on fluency, coherence, or factuality alone. Our benchmark introduces numeric/temporal consistency as an additional axis, particularly important when structured numeric and temporal data is translated into prose. This is inspired by both practical use cases (e.g., financial reporting) and gaps observed in current evaluation metrics. Our benchmarking not only evaluates the quality of the unstructured text that was generated, but we go one step further and provide a baseline for how we leverage the key-value extraction from this unstructured text to provide an additional layer of benchmarking of the concepts. This serves as a baseline key-value extractor that the community can build upon.


\section{Benchmark Generation} 
The benchmark generation can be broadly divided into three sub-tasks: (1) synthetic text generation from structured data, and (2) multi-dimensional quality assessment, (3) filtering low quality text with validation.

\subsection{Synthetic Text Generation}

The goal of our synthetic text generation pipeline is to produce ``natural'' language text from a tabular data source without the need for human intervention. Each row that is present in a table indicates a single instance of data, with the columns of the table indicating certain types of properties or relations of that data. 

Here, we use the term \textbf{reports} to refer to synthetic text generated for a particular set of columns of a table. A \textbf{report type} refers to all reports generated for a specified set of columns. The synthetic text generation step, then, will aim to produce one or more report types for each table and generate corresponding reports for each row. 

A report is more than just the task of converting the key-value pairs of structured data into templated text outputs but ensuring that the right combination of columns are selected to generate a report that transcends simple templating by dynamically determining column combinations and ensuring linguistic coherence.  

For the purposes of our benchmarking efforts, we assume that each row of the table will have its own corresponding set of synthetic texts. Given our ultimate goal of benchmarking and evaluating the quality of text-to-table extraction, in order to effectively produce data from which a system can feasibly extract the appropriate key-value information, we felt that the task would be unreasonably challenging for both extraction and evaluation if multiple rows could also be present within a given passage of text.

For a given table, we will produce several report types which target a set of columns within the table. For each report type, we then generate synthetic text for each row. Note that for a particular report type within a table, all rows will produce reports using the exact same set of columns, but for each table, a set of different columns will be selected.

We adopt a workflow that has a plan-then-execute two stage pipeline for our report generation.

\textbf{Planning Stage:} Our LLM based system first analyzes ten sample rows from the input table and must autonomously identify data patterns and relationships to produce meaningful report structures. Unlike structure based reporting which rely on predefined schema, the model analyzes column semantics, data types, and value distributions to understand the data's inherent structure. It must recognize natural groupings of related information, such as financial metrics paired with temporal context, and determine an appropriate level of granularity. We limit the output to one to five report types to ensure meaningful analytical aggregation rather than simple column enumeration. The LLM outputs a structured plan mapping each report type to its constituent elements.

\begin{figure}[h]
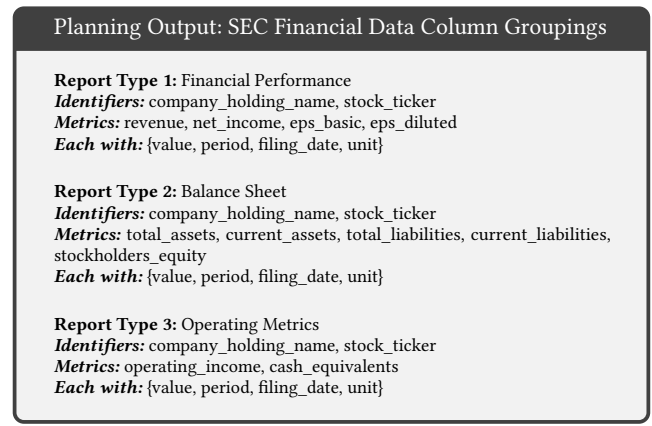

\centering
\begin{tcolorbox}[title=Planning Output: SEC Financial Data Column Groupings, colback=gray!10]
\footnotesize
\textbf{Report Type 1:} Financial Performance\\
\textbf{\textit{Identifiers:}} company\_holding\_name, stock\_ticker\\
\textbf{\textit{Metrics:}} revenue, net\_income, eps\_basic, eps\_diluted\\
\textbf{\textit{Each with:}} \{value, period, filing\_date, unit\}\\[0.3em]

\textbf{Report Type 2:} Balance Sheet\\
\textbf{\textit{Identifiers:}} company\_holding\_name, stock\_ticker\\
\textbf{\textit{Metrics:}} total\_assets, current\_assets, total\_liabilities, current\_liabilities, stockholders\_equity\\
\textbf{\textit{Each with:}} \{value, period, filing\_date, unit\}\\[0.3em]

\textbf{Report Type 3:} Operating Metrics\\
\textbf{\textit{Identifiers:}} company\_holding\_name, stock\_ticker\\
\textbf{\textit{Metrics:}} operating\_income, cash\_equivalents\\
\textbf{\textit{Each with:}} \{value, period, filing\_date, unit\}
\end{tcolorbox}
\caption{LLM planning phase output showing natural groupings of columns (metrics). Each report includes company identifiers and financial metrics with temporal/unit metadata.}
\label{fig:planning_example}
\end{figure}

For instance, when presented with SEC data, the model autonomously identifies that revenue, net income, and earnings per share naturally group into a financial performance report without being explicitly told that these represent income statement components (see Fig. \ref{fig:planning_example}). This planning phase directly tests the model's ability to recognize implicit data relationships and impose a meaningful structure on the raw tabular data. 

We place \emph{no hard constraint} on the set of columns per report type, the LLM is free to reuse any columns it deems semantically relevant. This is important for identifiers such as \texttt{company\_holding\_name} and
\texttt{stock\_ticker} that appear in every SEC report type and is useful as an identifier. We empirically chose 10 samples as a balance between computational efficiency and coverage; ablations on sample size remain as future work

\textbf{Generation Stage}: 
In the second stage, the grounded text generation uses the planned report type name and the subset of columns selected from the original structured data from stage one to produce a coherent narrative that strictly follows this identified set of columns. This way, the models must examine the data samples and determine appropriate reports based on natural column groundings, testing analytical judgments rather than template filling. Further, the generated prompts mitigate hallucination of information by focusing on the selected subset of columns from the data enforcing factual discipline during text creation. The design aims at a) enabling the LLM to recognize meaningful data relationships without human guidance, and b) generating informative text while maintaining factual discipline and minimizing hallucinations as much as possible. 

\begin{figure}[h]
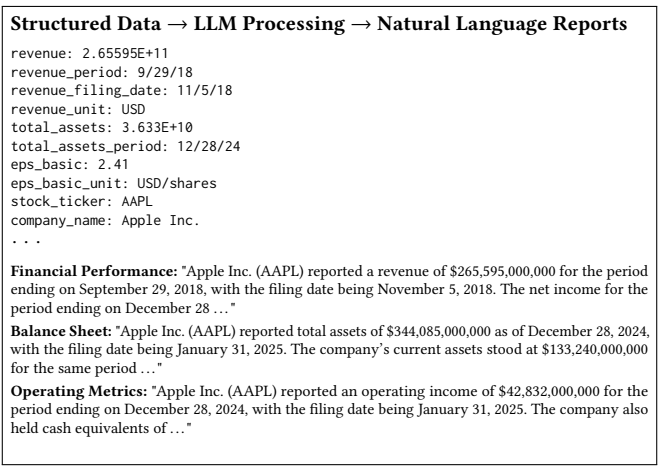

\centering
\small
\fbox{\parbox{\columnwidth}{
\textbf{Structured Data} → \textbf{LLM Processing} → \textbf{Natural Language Reports}\\[0.5em]
\begin{minipage}{0.28\textwidth}
\ttfamily\scriptsize
revenue: 2.65595E+11\\
revenue\_period: 9/29/18\\
revenue\_filing\_date: 11/5/18\\
revenue\_unit: USD\\
total\_assets: 3.633E+10\\
total\_assets\_period: 12/28/24\\
eps\_basic: 2.41\\
eps\_basic\_unit: USD/shares\\
stock\_ticker: AAPL\\
company\_name: Apple Inc.\\
\textbf{\dots} \\
\end{minipage}
\hfill
\begin{minipage}{\columnwidth}
\scriptsize
\textbf{Financial Performance:} "Apple Inc. (AAPL) reported a revenue of \$265,595,000,000 for the period ending on September 29, 2018, with the filing date being November 5, 2018. The net income for the period ending on December 28 \dots"\\[0.3em]
\textbf{Balance Sheet:} "Apple Inc. (AAPL) reported total assets of \$344,085,000,000 as of December 28, 2024, with the filing date being January 31, 2025. The company's current assets stood at \$133,240,000,000 for the same period \dots"\\[0.3em]
\textbf{Operating Metrics:} "Apple Inc. (AAPL) reported an operating income of \$42,832,000,000 for the period ending on December 28, 2024, with the filing date being January 31, 2025. The company also held cash equivalents of \dots"\\[0.3em]
\end{minipage}
}}
\caption{Example of LLM table-to-text transformation for Apple Inc. The model first plans report groupings, then generates natural language preserving exact values and temporal context.}
\label{fig:transformation_flow}
\end{figure}

\subsection{Multi-Dimensional Quality Assessment Framework}
To analyze the quality of the synthetically generated text, we employ a comprehensive validation framework which combines LLM-as-a-judge methods with objective extraction metrics to assess the generation quality and inspect the coverage and the faithfulness of information contained in the synthetic text.

\textbf{LLM-as-judge Evaluation:} 
We adopt the LLM as a judge paradigm~\cite{zheng-mtbench2020} to evaluate the quality of our generated reports along dimensions which are difficult to measure through other heuristic or objective measures. Specifically, we use LLMs to evaluate three critical aspects of the generated text: hallucination detection, where we aim to detect whether the generated text ``hallucinates'' any information which was not contained in the original structured data; coherence, which indicates how naturally the ideas in a text are presented; and factuality, or whether the generated text is factual from the perspective of general common sense knowledge. 

This evaluation methodology recognizes that the reports must be useful to human readers—not just technically accurate but coherent and free from hallucinations. For each measure, we design 5-point rubrics (See table \ref{tab:evaluation_rubrics}) to serve as the metric following a similar strategy as existing work (e.g., LLMs4Synthesis~\cite{10.1145/3677389.3702565}). This transforms the traditionally subjective human evaluation into a reproducible assessment.

The \textbf{hallucination} rubric identifies ungrounded content ranging from heavy fabrication of critical information (score 1) to complete grounding where all the content either is derived from the structured source data or includes explicit attribution (score 5). When judging the hallucination metric of a generated text, we ground the generated text based on the original tabular data that was used to generate the synthetic text -- note that generated text can be based on true information about the world but still count as a hallucination if the generated text contains information that goes beyond its grounding information.

The \textbf{coherence} rubric evaluates the narrative flow of the report generated and the logical organization of the different structures from incoherent texts with contradictions (score 1) to seamless organization with natural transitions (score 5). 

The \textbf{factuality} rubric assesses whether the claims in the general text are directly traced to the source data, distinguishing between fundamental incorrectness (score 1) (where most of the claims contradict) vs. full correctness (score 5).

These rubrics were designed to capture the nuances of the tabular data to synthetic text generation, wherein unlike in open-ended text generation, our rubrics emphasize verifiability, with each score level including concrete criteria for the source data relationships for both LLMs and humans enhancing reproducibility. Any automatic judge can itself hallucinate, and it's a known issue with using LLM-as-judge evaluations. However, we mitigate it by A) providing a scoring rubric to reduce the ambiguity of scores, and B) using self-consistency checks, where the LLM also produces a rationale for its score and an additional ``claims'' section, which judges each of the sentences in the report being judged to determine if the claim is supported or unsupported. These safeguards are inspired by the best practices in recent LLM-as-judge work
\cite{zheng2023mtbench,liu2023geval,jacovi2025facts}.

\begin{table*}[ht!]
\small
\caption{Evaluation rubrics for generated reports (1=worst, 5=best)}
\label{tab:evaluation_rubrics}
\begin{tabular}{p{0.8cm}p{5cm}p{5cm}p{5cm}}
\toprule
\textbf{Score} & \textbf{Factuality} & \textbf{Hallucination} & \textbf{Coherence} \\
\midrule
\textbf{1} & \textbf{Fundamentally incorrect:} Most claims contradicted & \textbf{Heavy:} Numerous invented details & \textbf{Incoherent:} Difficult to follow, random jumps \\
\textbf{2} & \textbf{Largely incorrect:} Core information misrepresented & \textbf{Frequent:} Multiple unverifiable points & \textbf{Poor flow:} Jarring transitions, disconnected \\
\textbf{3} & \textbf{Mixed accuracy:} Minor distortions, main narrative correct & \textbf{Occasional:} Some ungrounded details & \textbf{Acceptable:} Some awkward transitions \\
\textbf{4} & \textbf{Mostly correct:} Only minor/peripheral errors & \textbf{Rare:} Minor details lack grounding & \textbf{Smooth:} Clear progression, minor issues \\
\textbf{5} & \textbf{Fully correct:} No errors, all claims supported & \textbf{None:} All content grounded or attributed & \textbf{Seamless:} Natural flow, effortless transitions \\
\bottomrule
\end{tabular}
\end{table*}

\textbf{Numerical and Temporal Accuracy:} The numerical and temporal information in the synthetic report text is crucial to verify and requires special handling. If we generate synthetic text with incorrect numbers or dates (i.e., not aligned with the ground truth), it will be impossible for any downstream text-to-table approach to produce the correct extraction results. In order to validate the accuracy of such information in generated text, we implement a dedicated validation pipeline. 

For numeric values, we combine Stanford's Core NLP's NER~\cite{manning2014stanford, finkel2005incorporating} parser with regular expression patterns to identify monetary amounts, percentages, and quantities. We apply a 0.1\% relative error tolerance to account for the rounding differences. For each table row and associated synthetic text, we first collect the numeric cell values from the source ground truth row and apply the parser to produce normalised ground-truth numeric values. Next, for the generated text, we similarly apply the parser to identify all numeric values which were included in the text. Using the normalized results from these two parsers, we compare the two sets of numeric values to determine the precision and recall of numeric values in the generated text. We note that in this validation, we do not specifically check whether the numeric values are \textit{expressing the semantics} in the text accurately, but only check whether the numeric values occur.

Similarly, for temporal values, we employed a  LLM extraction~ for context-aware parsing of generated text, which handles complex phrases. 
Given that temporal identifiers such as dates and  times can be expressed in a large variety of ways -- e.g. ``the fourth quarter of 2022'' versus ``2022 Q4'' -- we used Stanford's Core NLP's SUTime~\cite{manning2014stanford} as a fallback to produce a standardized format. 

\subsection{Filtering Low Quality Text with Validation}

As a final step for the benchmark generation workflow, we can perform filtering to ensure that only high-quality data is used for evaluation purposes. 
Following the initial dataset collection and validation using the multidimensional quality assessment framework's results, we can analyze the quality assessment results to eliminate any reports that are not fully aligned with the ground truth structured data.

\textbf{Quality filtering}
After the multi-dimensional validation step, the filtering proceeds in three passes:
\begin{enumerate}[leftmargin=*]
\item \textbf{Metric selection}: pick the validation dimension whose metric is lowest for the dataset - e.g., we found that temporal accuracy was the lowest (see Table. \ref{tab:num_temp_eval_performance}).
\item \textbf{Thresholding}: drop any report whose \emph{precision or recall} on that dimension falls below a user-set threshold~$\tau$ (we sweep $\tau\!=\!1.0\!\rightarrow\!0.70$ in 0.05 steps; Fig.~\ref{fig:temporal-filtering} shows the trade-off).
\item \textbf{Re-weighting}: recompute the macro averages on the retained data. This helps validate if the improvements from pruning one metric adversely affects the others. 
\end{enumerate}
This simple strategy kept over 50\% of SEC reports when $\tau\!=\!0.90$ while also improving the overall metrics.

\section{Text-To-Table Extraction Baseline} 

Having established our methodology for generating synthetic text from tabular data, we now describe our baseline workflow for the text-to-table extraction task. Here, the main goal of the text-to-table extraction task is to take a collection of natural language texts and convert it into a tabular data format, with each piece of text corresponding to a single row with one or more columns and values.

\subsection{Baseline Extraction Approach}

For the scope of this paper and benchmark creation, we make two key assumptions in this task. First, we assume that a given piece of text corresponds to only a single row, rather than requiring extraction of multiple rows from a single chunk of text. Second, for the extraction and evaluation steps, we assume that in a collection of texts, a consistent set of columns can be extracted from each text. Relating this back to our automatic report generation workflow, each report corresponds to a set of columns from the table, and each report type is constructed using a fixed set of columns. Our assumption follows from this structure, where our extraction and evaluation is performed for a specific report type. This assumption allows us to evaluate two aspects of a text-to-table extraction method -- the ability to correctly identify what column(s) can be extracted from a set of report texts, and the ability to correctly extract values.

Our extraction approach operates over a single report type at a time for each table. The first step is to inspect a sample of the reports to identify a set of columns which can be extracted from the reports. Not all reports might necessarily contain information about all columns (e.g., this can occur if the original row had null values for a given column), but by sampling over multiple reports our system will aim to get a general sense of the kind of information which can be extracted. Here, we make use of an LLM prompting strategy to identify a set of plausible columns which can be extracted from the report type. 

Next, given the set of columns which we have predicted can be extracted from the reports, we perform extraction over each row's report text. For each row, we feed in the report text together with the predicted set of columns to extract, and produce a JSON-formatted dictionary of key-value pairs corresponding to column-value extractions. 

Together, these two steps compose our baseline implementation for extracting tables from text, with the first step essentially extracting the ``schema'' of the table and the second extracting the cell values. For tables where multiple reports were generated, we repeat this process in a similar manner -- for the scope of this paper, we process each report independently and focus only on evaluating the extraction quality for each report in isolation.

\subsection{Evaluating Extraction Quality}

In order to evaluate the quality of the text-to-table extraction results, there are two factors which must be considered. First, it is possible that extraction will identify column names which are not identical matches with the original column names. In such cases we will need to rely on other similarity metrics to avoid overly penalizing the results for minor differences such as singular vs plural, tense, capitalization, or spacing differences. Besides checking for exact matches, we can use edit-based similarity metrics such as Levenshtein distance~\cite{levenshtein1966binary}, similar to the work of KVP10K~\cite{NaparstekASAYBPRDPFALLN24}, or embedding-based similarity such as BERTScore~\cite{bert-score}. This flexibility allows our baseline approach to prioritize semantically similar column names over exact matches helping adapt to a wide range of naming outcomes in real-world data. Ultimately, the focus is to unify values under appropriate columns- irrespective of the exact name as long as the mappings preserve the correct semantics and the extracted values align with the ground truth.

A second consideration is how to determine which predicted column extraction to compare against the ground truth columns. While this might be trivial in cases where the extraction has exact matches, when there are minor differences in the column names we must make a decision about which extracted column values will be compared against which ground truth columns. Furthermore, this decision may be further complicated when handling cases where the system predicts too many or too few columns to extract from each report. 

To handle this, we employ a bipartite matching approach, mapping each ground truth column to at most one predicted column. Each pair of ground truth and predicted column will have some associated similarity score, computed using the aforementioned similarity metrics. We then choose the best mapping by selecting the mapping which maximizes the total similarity scores among the mapped columns. We apply OR tools~\cite{ortools} to optimize the total score using integer programming.

Lastly, after performing extraction and choosing optimal mappings to compare predictions against ground truth columns, we can evaluate the extraction quality. We apply standard measures of precision, recall, and F1 scores, weighting each metric by the text similarity score. For our text-to-table extraction task, we can evaluate how accurately we identify columns (for each report of each table) as well as the accuracy of value extractions (for each row of each report).

\section{Experimental Methods}
\subsection{Datasets}
We evaluate our benchmark generation method across datasets from various diverse domains representing real-world complexity: 

\textbf{SEC Financial Filings:} The dataset consists of structured financial data extracted from SEC 10-K and 10-Q filings~\cite{sec}. This dataset was programmatically extracted from the SEC Edgar filings via the official API. Synthetic text generated from SEC Financial Filings data offers significant advantages, wherein the researchers can reproduce and extend the dataset without data distribution concerns, and the extraction process itself serves as a realistic structured table-to-text benchmark.

Creation pipeline retrieves company facts from the SEC's XBRL API~\footnote{\url{https://www.sec.gov/search-filings/edgar-application-programming-interfaces}} for publicly traded companies for which the ticker symbol has the relevant information that could be extracted through the API. We extract the financial metrics by searching for the most commonly used US-GAAP taxonomy~\cite{de2000xbrl} concepts, including multiple naming variations to maximize coverage (e.g., revenue, revenues, and sales revenue net all map to revenue). Our selection criteria focused on metrics universally present in financial analysis (income statement items, balance sheet components, and cash positions). Each metric includes temporal metadata and unit information, resulting in over 40 columns per company. 

This approach ensures that the benchmark can be regenerated using the provided extraction code while capturing the complexity of real financial reporting (where the same concept may have multiple representations and temporal context). The code handles both annual and quarterly filing, prioritizing the most recent data when multiple values exist. 

\begin{figure*}[h]
    \includegraphics[width=0.8\textwidth]{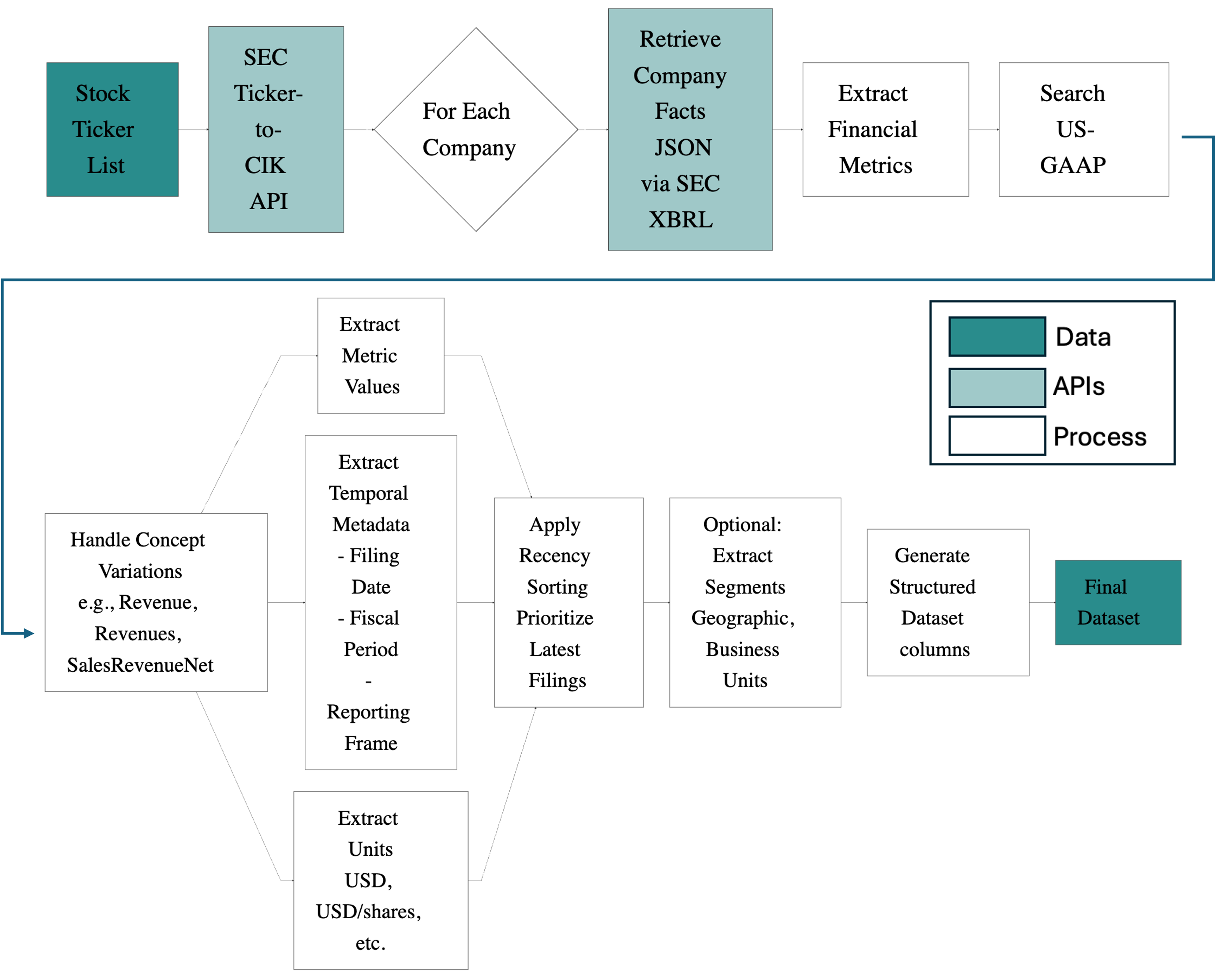}
    \caption{SEC dataset extraction pipeline showing the programmatic generation of structured financial data from EDGAR filings. The process handles concept variations and temporal complexity inherent in financial reporting.}
    
    \label{fig:enter-label}
\end{figure*}

\textbf{WikiDBs:} WikiDBs~\cite{vogel2024wikidbs} is a dataset that is derived from Wikidata~\cite{WikiData}, where knowledge graph entities and their relations have been curated and formatted to resemble tabular data. WikiDBs leverages the large-scale and well-structured contents of Wikidata to collect large groups of entities with related topics and properties and then transforms the assertions present in the KG into columns and cell values. Like Wikidata itself, WikiDB covers an extremely broad range of topics, and the full dataset contains 100,000 tables. 

For our initial experiments, we collected a subset of 1,000 tables from WikiDBs to validate our approach, with the included tables covering a diverse set of topics such as geographic data, cultural and economic indicators, product and service information, and legislative data. Additionally, we filter out any columns containing potential personally identifiable information, such as names and contact information. The final version of the tables contained an average of 11.6 columns and 152.8 rows.

\subsection{Models and Infrastructure}

We used the Qwen2.5-72B-Instruct model~\cite{qwen2.5}, which provided the report planning and generation. For the evaluation part, we used the Meta's Llama-3.3-70B-Instruct~\cite{grattafiori2024llama3herdmodels} as our primary LLM for all the evaluation components. We set the temperature to zero for deterministic generation and employed the DSPy~\cite{dspy} framework to enforce the structured outputs processing averages.

\subsection{Evaluation Protocol} 

The final scores that were combined using the multiple evaluation perspectives. The generated text quality averages across the three LLM-as-judge dimensions while the information fidelity weighs the column identification and the value extraction quality. The data set level application uses size weighted averages to prevent the small data sets from disproportionately influencing the overall conclusion. 

\section{Results}
We evaluated our benchmark on 50 datasets spanning SEC's financial filings and a subset of diverse WikiDB tables. Our multi-dimensional evaluation framework assesses text quality, information fidelity, and numeric and temporal accuracy to provide a comprehensive view of the table-to-text and generation quality.

\subsection{LLM-as-a-Judge Evaluation Results}

Table \ref{tab:text_quality_comparison} presents the LLM-as-judge evaluation results across three dimensions with factuality, hallucination detection, and coherence.  The results demonstrate exceptionally strong performance in factual accuracy with scores of 4.58 \& 4.56 for factuality and 4.90 \& 4.55 for hallucination avoidance. 

These near ceiling scores indicate that the current LLMs excel at generating accurate content without introducing fabricated information, and the way the two-step structured text to report generation seems to be effective. However, coherence scores reveals a notable performance gap representing the challenges of having a coherent text. 

Interestingly, both the SEC financial data and the WikiDB show markedly different characteristics. The SEC reports achieve near perfect hallucination avoidance but struggle with coherence, while the WikiDB shows more balanced performance across dimensions. These patterns suggest that the model has effectively extracted and conveyed the factual information from such resources but face systematic challenges in organizing this information into naturally flowing narratives.

\subsection{Numerical and Temporal Accuracy}

Having accurate numerical and temporal information is critical for our benchmark creation, as it would be impossible for any text-to-table approach to extract the correct information if it is not included in our generated reports. Our temporal and unit accuracy evaluation reveals strong performances in preserving specific factual details. Table \ref{tab:num_temp_eval_performance} presents the precision, recall, and F1 scores for numeric and temporal information validation in the generated reports. The results demonstrate strong performances with both numeric extractions and temporal extraction achieving a score greater than 0.9. These scores represent the highest performance across all evaluation dimensions, suggesting that the models excel at preserving specific values.

Across the different metrics, the models have successfully identified and included nearly all the numeric and temporal information from the source data. The slightly lower precision suggests occasional confusion between similar values or dates, 
where different granularities and contextual expressions create ambiguity in what constitutes a "correct" extraction. The strong performance contrasts to the key-value extraction highlighting that while specific facts are preserved accurately, the semantic relations and attributions become obscured in narrative form.

\subsection{Baseline Text-To-Table extraction Results}
To assess the difficulty of the task for an LLM and also to enable more advanced systems to compare with some baseline results, we evaluated the column and value extraction performance with our baseline implementation. To compare against the ground truth columns and values, we use a normalized Levenshtein edit distance, and compute the similarity of predicted and true values as 1 minus the edit distance. Table \ref{tab:extraction_performance} shows both column identification and value extraction tasks' precision, recall, and F1 scores. The results revealed significant headroom in performance in automated information extraction from generated text, demonstrating that the benchmark is sufficiently challenging for LLMs indicating that while the information being present in the native text, they are expressed in ways that make automated extraction difficult. The precision-recall trade-offs between them differ notably between the different dataset types. The SEC dataset shows higher recall but very low precision, while WikiDBs achieve a more balanced performance. This pattern suggests that a fundamental challenge for SEC data is that it tends to include more information from source data, with complexities in how it's expressed. At the same time, WikiDB reports are more selective or slightly more precise.

\begin{table}[htbp]
\caption{Text Quality Evaluation (LLM-as-judge) scored from 1=worst to 5=best}
\label{tab:text_quality_comparison}
\resizebox{\columnwidth}{!}{%
\begin{tabular}{@{}lrcccc@{}}
\toprule
\textbf{Dataset Type} &\textbf{Factuality} & \textbf{Hallucination} & \textbf{Coherence} & \textbf{Overall} \\
\midrule
SEC Financial & 4.58 & 4.90  & 3.28  & 4.25 \\
WikiDB Tables & 4.56 & 4.55  & 3.53  & 4.21 \\
\bottomrule
\end{tabular}
}
\end{table}

\begin{table}[htbp]
\centering
\caption{Text Quality Evaluation (Numeric and Temporal Accuracy)}
\label{tab:num_temp_eval_performance}
\resizebox{\columnwidth}{!}{%
\begin{tabular}{@{}llccc@{}}
\toprule
\textbf{Validation Type} & \textbf{Dataset} & \textbf{Precision} & \textbf{Recall} & \textbf{F1} \\
\midrule
\multirow{3}{*}{Numeric Validation} & SEC Financial & 0.941 & 0.924 & 0.927 \\
 & WikiDB Tables & 0.849 & 0.971 & 0.956 \\
\midrule
\multirow{3}{*}{Temporal Validation} & SEC Financial & 0.818 & 0.977 & 0.956 \\
 & WikiDB Tables & 0.878 & 0.916 & 0.915 \\
\bottomrule
\end{tabular}
}
\end{table}

\begin{table}[htbp]
\centering
\caption{Extraction Performance Across Information Types}
\label{tab:extraction_performance}
\resizebox{\columnwidth}{!}{%
\begin{tabular}{@{}llccc@{}}
\toprule
\textbf{Extraction Type} & \textbf{Dataset} & \textbf{Precision} & \textbf{Recall} & \textbf{F1} \\
\midrule
\multirow{3}{*}{Column Identification} & SEC Financial & 0.344 & 0.669 & 0.455 \\
 & WikiDB Tables & 0.395 & 0.433 & 0.413 \\
\midrule
\multirow{3}{*}{Value Extraction} & SEC Financial & 0.257 & 0.11 & -- \\
 & WikiDB Tables & 0.179 & 0.137 & -- \\
\bottomrule
\end{tabular}
}
\end{table}

\begin{figure*}[h]
    \includegraphics[width=\textwidth]{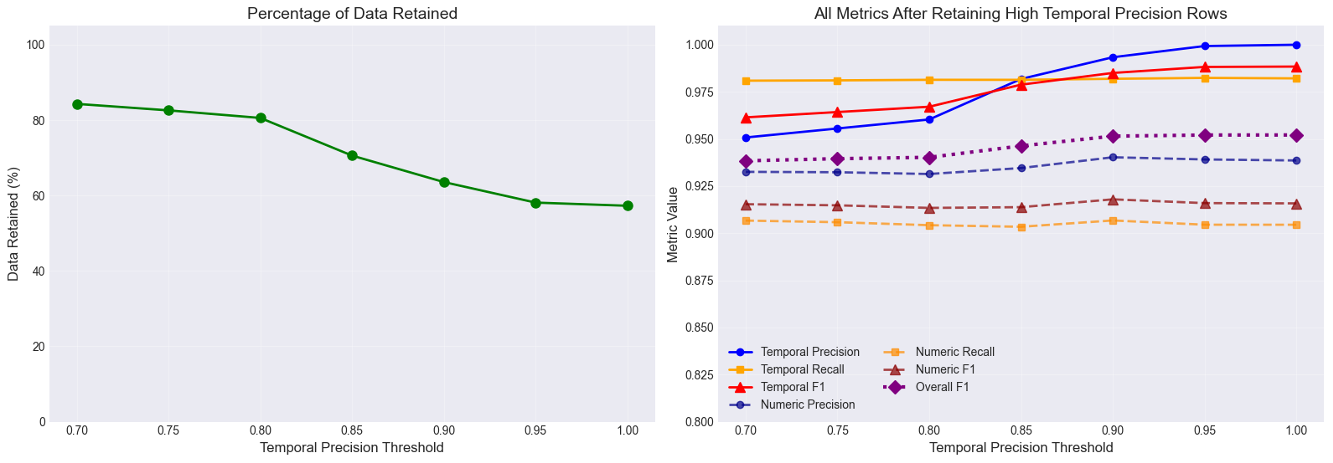}
    \caption{Impact of filtering on the SEC dataset, showing the percentage of total data remaining after applying filtering with various threshold values~$\tau$ (left) and the change in quality metrics over the remaining data as filtering is applied (right).}
    \label{fig:temporal-filtering}
\end{figure*}

\newlength{\figboxwd}
\setlength{\figboxwd}{\dimexpr\columnwidth-2\fboxsep-2\fboxrule\relax}

\begin{figure}[t]
\centering
\small
\fbox{%
\begin{minipage}{\figboxwd}
\ttfamily\scriptsize\raggedright\setlength{\emergencystretch}{1em}

\textbf{Generated Text}: Hilton Worldwide Holdings Inc. (stock ticker: HLT) reported an operating income of \$2{,}370{,}000{,}000 for the period ending 12/31/24, with the filing date on 2/6/25. The company's cash and cash equivalents were \$1{,}301{,}000{,}000 as of 12/31/24, also filed on 2/6/25. These operating metrics highlight the company's operational efficiency and liquidity position at the end of the fiscal year \textcolor{red}{2024}.

\medskip\noindent\rule{\linewidth}{0.4pt}\medskip

\textbf{Number of extracted dates using the temporal parser}: 5\\
\textbf{Extracted Dates from Text}: %
\begin{tabularx}{\linewidth}{@{}X@{}}
[{\{'value': `2024-12-31', `type': `date', `text': `12/31/24'\}, \{'value': `2025-02-06', `type': `date', `text': `2/6/25'\}, \{'value': `2024-12-31', `type': `date', `text': `12/31/24'\}, \{'value': `2025-02-06', `type': `date', `text': `2/6/25'\}, \textcolor{red}{\textbf{\{'value': `2024', `type': `year', `text': `fiscal year 2024'\}}}}]
\end{tabularx}

\medskip\noindent\rule{\linewidth}{0.4pt}\medskip

{\renewcommand{\arraystretch}{0.9}
\textbf{Ground Truth Date Columns with the Parsed Dates:}\\ 
\begin{tabularx}{\linewidth}{@{}l >{\raggedright\arraybackslash}X l@{}}
\toprule
\textbf{Column} & \textbf{Cell Value} & \textbf{Parsed} \\
\midrule
stock\_ticker                   & HLT       & [] \\
operating\_income\_period       & 12/31/24  & 2024-12-31 \\
operating\_income\_filing\_date & 2/6/25    & 2025-02-06 \\
cash\_equivalents\_period       & 12/31/24  & 2024-12-31 \\
cash\_equivalents\_filing\_date & 2/6/25    & 2025-02-06 \\
\bottomrule
\end{tabularx}}

\medskip

\textbf{Temporal Parsing Metrics:}\\[-2pt]
\begin{tabular}{@{}ll@{}}
Precision:       & 80\% (4 correct / 5 extracted) \\
Recall:          & 100\% (4 found / 4 ground truth) \\
F1 Score:        & 88.89\% \\
True Positives:  & 4 \\
False Positives: & 1 \\
False Negatives: & 0 \\
\end{tabular}

\end{minipage}}
\caption{
Parsing Challenges: The temporal metrics reveal a precision of 80.00\% and recall of 100\%, indicating that while all ground truth dates were captured, the parser extracted additional temporal expressions. The lower precision is attributed to the LLM correctly identifying "2024" from "fiscal year 2024" (shown in bold), which represents valid temporal information but was not included in the ground truth annotations. This highlights the inherent challenges in temporal parsing where different granularities (full dates vs. years) and contextual expressions create ambiguity in what constitutes a "correct" extraction.
}
\Description{Example figure showing generated text, extracted dates, a table of ground truth dates versus parsed dates, and evaluation metrics.}
\label{fig:parsing_challenge}
\end{figure}


\subsection{Quality Filtering}

To investigate the impact of performing quality filtering over the generated benchmark data, we also performed a granular analysis of the numeric and temporal accuracy. Temporal precision emerged as the weakest component (see Table \ref{tab:extraction_performance}), so we applied a threshold filter over this metric and evaluated the dataset retention and performance. We inspect the impact of threshold values~$\tau$ across temporal precision at intervals of 0.05 starting from $\tau=1.0$ down to $\tau=0.7$, and quantify how filtering impacts dataset size and metric performance. Fig \ref{fig:temporal-filtering} shows the performance in the SEC dataset, which appeared to be particularly challenging for our report generation approach. Despite aggressive filtering, over 50\% of the dataset was retained. 

Figure \ref{fig:parsing_challenge} shows the challenges of validating the numeric and temporal accuracy. Although the LLM generated text correctly identified the fiscal year as 2024 as a temporal value, this was not explicitly contained within the ground truth columns. This highlights the challenges of parsing the different contextual expressions and granularities of data -- neither the text generation nor quality evaluation were technically incorrect in their outputs, but taken as a whole the lowered precision of this example is misleading.

\section{Discussion}
\subsection{Key Findings and Contributions}

While the models excelled at factual accuracy and minimizing hallucinations they struggled with narrative coherence and information extractability. The dichotomy highlights a fundamental challenge: generating text that is both accurate and useful requires not just correct fact but also effective organization and clear attribution.

While strong performance on numerical and temporal accuracies demonstrated that the models can preserve specific numerical and the temporal details effectively, the poor key-value extraction results suggest that these facts become embedded in narrative structures and obscure the semantic relations to source columns. This finding has important implications for practical applications where downstream systems need to extract such information from generated reports. We also note that the method that we produced is a baseline to give the community a benchmark to build upon and recognize that this is by no means the best method out there for text-to-table extraction methodologies.

\subsection{Scalability and Extensibility}
 While our evaluation focused on a representative subset of 50 tables from the WikiDBs dataset, our dataset collection pipeline generated significantly larger resources. We also share a larger subset of WikiDBs, formatted to be suitable for easy consumption and evaluation through our workflow, for the community to further build upon. 
 
 For the SEC dataset, in our current experiments we limited it to only include US-GAAP taxonomy concepts, but the same code for dataset creation can be extended to encompass the full XBRL taxonomy system, potentially increasing coverage by an order of magnitude.

 Beyond the datasets we release to support experimentation and validation of our methodology, our benchmark generation workflow is extensible to any new tabular dataset. We hope that our approach can help to greatly expand the scope of benchmark creation and evaluation for the text-to-table task.

\subsection{Tools and Community Infrastructure}

In our use of LLM as judge we designed the prompt rubric, we tried to keep it as general as possible. Further enhancement could improve context-aware rubrics that could be tailored to specific industries depending on the use case to further enhance the requirements and trying to match the judgment. Another future work is to do inter-human and inter-model judge capability assessment to see how close or how far apart the two are when provided with similar instructions for evaluating the different metrics. 

Our numeric-temporal extraction pipeline, while achieving strong results, represents a first-pass solution to a complex problem. Further improvements could include context-aware, numerical normalization, handling multiple relative temporal expressions, and multi-scale temporal reasoning (such as quarters, years, and filing dates). 

We positioned our key-value extraction system as the baseline. Substantial room exists for improvement. We encourage the community to develop enhanced extraction methods that better handle semantic paraphrasing and distributed information and implicit relationships between values and their attributes.

Our quantitative study focused on a single strong open-weights model (Qwen2.5-72B-Instruct model for report generation and Meta's Llama-3.3-70B-Instruct for evaluation) so that we could prototype the end-to-end evaluation pipeline and demonstrate our benchmarking and evaluation methods. The overall pipeline, however, is model-agnostic and can be used with any combination of models. In future work, we plan to benchmark a spectrum of model sizes and architectures. We release the generation and evaluation code to encourage the community to run—and publish—such cross-model comparisons.

\section{Conclusion}
This paper establishes a comprehensive workflow and evaluation infrastructure to generate benchmarks for text-to-table extraction tasks, addressing a critical gap between research evaluation and practical needs. Our contribution spans three key areas: 

\begin{itemize}
    \item We provide a large-scale, multi-dimensional dataset with realistic complexity. Unlike existing benchmarks that focus on narrow tasks or simplified scenarios, our data reflects the challenges of real-world reporting systems where accuracy, completeness, and readability must be balanced.
    \item Our multi-dimensional evaluation framework goes beyond surface-level metrics. Our combination of LLM as judge assessment with objective extraction metrics will reveal that current models can generate factually accurate text that remains difficult to process programmatically, finding with important implications for system design.
    \item We deliver complete open source infrastructure enabling reproducible research and community extension. Our baseline tools, while showing room for improvement, provide concrete starting points for advancing the field. This benchmark reveals both achievements and challenges in current LLM capabilities. 
\end{itemize}

While models demonstrate strong factual grounding, numerical precision, and hallucination mitigation, they struggle with narrative organization and semantic clarity. These findings suggest that the future research should focus not just on accuracy but on generating texts that support downstream information extraction. By providing a comprehensive benchmark, we aim to accelerate the progress in table-to-text generation towards systems that meet the requirements of real-world applications.


\bibliographystyle{ACM-Reference-Format}
\bibliography{refs}

\end{document}